    \documentclass[10pt,twocolumn,letterpaper]{article}
    
    \usepackage{cvpr}              
    
    \usepackage{graphicx}
    \usepackage{amsmath}
    \usepackage{amssymb}
    \usepackage{booktabs}
    
    \usepackage[utf8]{inputenc} 
    \usepackage[T1]{fontenc}    
    \usepackage{url}            
    \usepackage{booktabs}       
    \usepackage{amsfonts}       
    \usepackage{nicefrac}       
    \usepackage{microtype}      
    \usepackage{graphicx}
    \usepackage{booktabs}
    \usepackage[normalem]{ulem}
    \usepackage{multirow}
    \usepackage{amsmath}
    \usepackage{makecell}
    \usepackage{stfloats}
    \usepackage{wrapfig}
    \usepackage{enumitem}
    \usepackage{scalerel}
    \usepackage{color}
    \usepackage[dvipsnames]{xcolor}
    
    \usepackage{algorithm}
    \usepackage{algorithmic}

    \DeclareMathOperator*{\argmin}{argmin}
    
    \usepackage[pagebackref,breaklinks,colorlinks]{hyperref}

    \usepackage[capitalize]{cleveref}
    \crefname{section}{Sec.}{Secs.}
    \Crefname{section}{Section}{Sections}
    \Crefname{table}{Table}{Tables}
    \crefname{table}{Tab.}{Tabs.}

    \def\cvprPaperID{8777} 
    \def\confName{CVPR}
    \def\confYear{2023}

\begin{document}
    
    \title{Long-term Visual Localization with Mobile Sensors}
    
    \author{
        Shen Yan$^{1,3}$ 
        \quad Yu Liu$^{1}$ 
        \quad Long Wang$^{2}$ 
        \quad Zehong Shen$^{3}$ 
        \quad Zhen Peng$^{2}$\\
        \quad Haomin Liu$^{2}$
        \quad Maojun Zhang$^{1}$
        \quad Guofeng Zhang$^{3}$
        \quad Xiaowei Zhou$^{3\dagger}$\\[1.5mm]
        $^1$National University of Defense Technology\quad
        $^2$SenseTime Research\quad
        $^3$Zhejiang University \quad 
    }

    \maketitle

    \begin{abstract}
    Despite the remarkable advances in image matching and pose estimation, image-based localization of a camera in a temporally-varying outdoor environment is still a challenging problem due to huge appearance disparity between query and reference images caused by illumination, seasonal and structural changes. In this work, we propose to leverage additional sensors on a mobile phone, mainly GPS, compass, and gravity sensor, to solve this challenging problem. We show that these mobile sensors provide decent initial poses and effective constraints to reduce the searching space in image matching and final pose estimation. With the initial pose, we are also able to devise a direct 2D-3D matching network to efficiently establish 2D-3D correspondences instead of tedious 2D-2D matching in existing systems.
    As no public dataset exists for the studied problem, we collect a new dataset that provides a variety of mobile sensor data and significant scene appearance variations, and develop a system to acquire ground-truth poses for query images. 
    We benchmark our method as well as several state-of-the-art baselines and demonstrate the effectiveness of the proposed approach. The code and dataset will be released publicly.
    \end{abstract}

    \let\thefootnote\relax\footnotetext{The authors from Zhejiang University are affiliated with the State Key Lab of CAD\&CG and ZJU-SenseTime Joint Lab of 3D Vision. $^\dagger$Corresponding author: Xiaowei Zhou.}

     
    \section{Introduction}
    \label{sec:intro}
    
    Visual localization aims at estimating the camera translation and orientation for a given image relative to a known scene. Solving this problem is crucial for many applications such as autonomous driving~\cite{cheng2019cascaded}, robot navigation~\cite{lim2012real} and augmented and virtual reality~\cite{castle2008video, lynen2015get}.
    
    State-of-the-art approaches to visual localization typically involve matching 3D points in a pre-built map and 2D pixels in a query image~\cite{brachmann2021visual,germain2020s2dnet,lynen2020large,sarlin2019coarse,sarlin2020superglue,sattler2016efficient,taira2018inloc,toft2020long}. An intermediate image retrieval step~\cite{arandjelovic2016netvlad, revaud2019learning, gordo2017end, ge2020self} is often applied to determine which parts of the scene are likely visible in the query image, in order to handle large-scale scenes. The resulting camera poses are estimated using a standard Perspective-n-Point (PnP) solver~\cite{haralick1994review, kneip2011novel} inside a robust RANSAC~\cite{barath2019magsac, chum2008optimal, fischler1981random, chum2003locally} loop. However, in real-world outdoor scenarios, obtaining such correspondences and further recovering the 6-DoF pose are difficult, since the outdoor scenes can experience large appearance variations caused by illumination (e.g., day and night), seasonal (e.g., summer and winter) and structure changes. Visual localization under such challenging conditions remains an unsolved problem, as reported by recent benchmarks~\cite{sarlin2022lamar, sattler2018benchmarking, zhang2021reference}.  
    
    %
    
    \begin{figure}[t]
       \centering
       \includegraphics[width=0.98\linewidth]{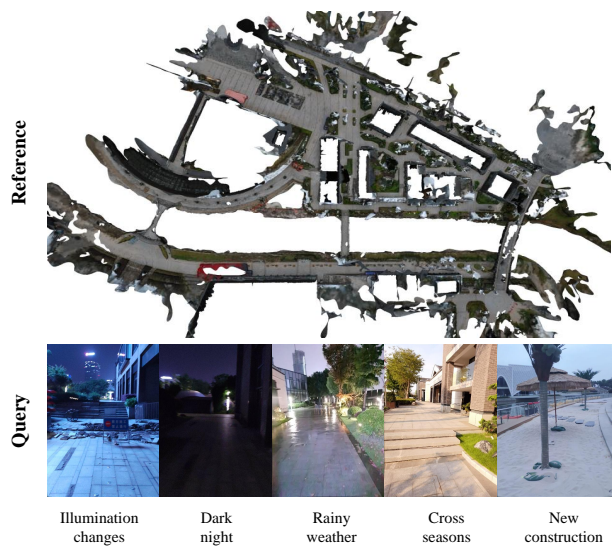}
    
       \caption{\textbf{Visual localization under extremely challenging conditions.} The proposed benchmark dataset \textit{SensLoc} exhibits long-term appearance changes due to illumination, weather, season, day-night alternation, and new constructions.}
       \label{fig:dataset1}
    \end{figure}    

    Fortunately, nowadays, with the popularity of smart devices that come equipped with various sensors such as Inertial Measurement Unit (IMU), gravity, compass, GPS, or radio signals (like WiFi and Bluetooth), new possibilities arise for mobile phone pose estimation exploiting these additional multi-modality sensors. Nevertheless, previous works only take independent sensors into consideration. For example, some methods utilize GPS as a prior to bound the Visual-Inertial Odometry (VIO) drift~\cite{yu2019gps, schreiber2016vehicle, kim2014high, qin2019general} or simplify the image retrieval process~\cite{li2018integration, vishal2015accurate, vysotska2015efficient}, while others focus on employing the gravity direction as a reliable prior to benefit the PnP solver~\cite{albl2016rolling, svarm2016city, sattler2016efficient, sweeney2015efficient, fragoso2020gdls, kukelova2010closed, svarm2014accurate}.
    
    In this paper, we introduce a novel framework, named SensLoc, that localizes an image by tightly coupling visual perception with complementary mobile sensor information for robust localization under extreme conditions. In the first stage, our approach leverages GPS and compass to constrain the search space for image retrieval, which not only reduces the risk of misrepresentation of global features but also speedups the retrieval procedure with fewer database candidates. In the second stage, inspired by the recent CAD-free object pose estimation method, OnePose++~\cite{he2022onepose}, we design a transformer-based network to directly match 3D points of the retrieved sub-map to dense 2D pixels of the query photo in a coarse-to-fine manner. Compared to the modern visual localization pipeline, which establishes 2D-3D correspondences by repeatedly matching 2D-2D local features between query and retrieve images, our solution shows a significant acceleration and a better performance especially under challenging appearance changes. In the last stage, we implement a simple yet effective gravity validation algorithm and integrate it into the RANSAC loop to filter the wrong pose hypotheses, leveraging the precise roll and pitch angles from mobile gravity sensors. The gravity validation leads to an improvement of RANSAC in terms of efficiency and accuracy, as false hypotheses can be removed in advance.
    

    To the best of our knowledge, there is no public dataset for multi-sensor localization under strong visual changes. To facilitate the research of this area, we created a new benchmark dataset \textit{SensLoc}, as shown in \cref{fig:dataset1}. Specifically, we first used a consumer-grade panoramic camera (Insta360) and a handheld Real-time Kinematic (RTK) recorder, to capture and reconstruct a large-scale reference map. Half a year later, we collected query sequences with large scene appearance changes through a mobile phone bounded with a RTK and recorded all available built-in sensor data. 
    As direct registration between the query sequences and the reference map is difficult,
    we rebuilt an auxiliary map at the same time as acquiring the query sequences using Insta360 and RTK and aligned the auxiliary map with the reference map through ICP. 
    Thus, we only needed to register the query images with the auxiliary map, which was easier as they were captured at the same time. 
    To achieve this, we developed a pseudo-ground-truth (GT) generation algorithm to accurately register each query sequence against the auxiliary map by incorporating feature matching, visual-inertial-odometry, RTK positions and gravity directions. The GT generation algorithm does not ask for any manual intervention or extra setup in the environment, enabling scalable pose labeling.
    

    We evaluate several state-of-the-art image retrieval and localization methods on our proposed dataset. We show that the performance of the existing methods can be drastically improved by considering sensor priors available in mobile devices, such as GPS, compass, and gravity direction. The experiments also demonstrate that our method outperforms the state-of-the-art approach HLoc~\cite{sarlin2019coarse, sarlin2020superglue} by a large margin in challenging night-time environments, while taking only $66ms$ to find 2D-3D correspondences on a GPU and $8ms$ for PnP RANSAC. 
    
    In summary, our main contributions include:
    \begin{itemize}
        \item A novel outdoor visual localization framework with multi-sensor prior for robust and accurate localization under extreme visual changes.
        \item A new dataset for multi-sensor visual localization with seasonal and illumination variations.
        \item Benchmarking existing methods and demonstrating the effectiveness of the proposed approach.
    \end{itemize}
    
    
     \begin{table*}[t]
    \centering
    \resizebox{0.9\textwidth}{!}{
    \setlength\tabcolsep{2pt} 
    	\begin{tabular}{l||ccccccc} 
    		\Xhline{3\arrayrulewidth}
        	dataset  & scale & equipment cost & operation cost & changes & additional sensors & groundtruth solution & accuracy \\ 
    		\hline
    		Cambridge~\cite{kendall2015posenet} & small & low  & low & people,weather & No & SfM & $\textgreater dm$      
    		\\
    		Phototourism~\cite{jin2021image} & small & low & low & people,construction & No & SfM & $\approx m$                                               \\
    		San Francisco~\cite{chen2011city} & large & low & low & people,construction & GPS & SfM+GPS & $\approx m$     
                                                            \\
    		Aachen~\cite{sattler2018benchmarking} &  large & low & high & people,day-night & No & SfM+manual & $\textgreater dm$                                                \\
    		NCLT~\cite{carlevaris2016university} & median & high & low & weather & LiDAR, IMU, GPS & VIO+LiDAR & $\approx dm$                                                           \\
    		ADVIO~\cite{cortes2018advio}  & median & median & high & people & IMU, depth, GPS & VIO+manual & $\approx m$
    		                                \\
    		ETH3D~\cite{schops2017multi} & small & high & high & No & LiDAR & LiDAR+manual & $\approx mm$
    		                                \\         
    		\multirow{2}*{LaMAR~\cite{sarlin2022lamar}} & \multirow{2}*{large} & \multirow{2}*{high}  & \multirow{2}*{low} & people,weather, & LiDAR, IMU, WiFi, & LiDAR+ & 	\multirow{2}*{$\approx cm$}
    		                                \\  
    		&&&& day-night,construction & Bluetooth, depth, infrared & SL+VIO
    		\\
    		\hline       
    		\multirow{2}*{SensLoc} & \multirow{2}*{large} & \multirow{2}*{low} & \multirow{2}*{low} & people,car,weather, & GPS, IMU, WiFi, & SL+VIO+ & \multirow{2}*{$ \textless dm$} 
    		                                \\ 
    		&&&& day-night,construction & Bluetooth, compass, gravity & RTK+Gravity
    		                                \\
    		\Xhline{3\arrayrulewidth}
    	\end{tabular}
    }
    \caption{\textbf{Overview of existing outdoor datasets.} The following attributes are taken in account: environment (scale, changes), cost (equipment, operation), groundtruth (solution, accuracy), and whether contains sensor data. The equipment and operation costs refer to capital and labor overhead when building the dataset.}
    \label{tab:dataset}
 \end{table*}

 \section{Related work}
    \paragraph{Visual Localization.}
    Conventionally, most existing methods~\cite{sarlin2019coarse,sarlin2020superglue,lowe2004distinctive,sun2021loftr} tackle pose estimation by establishing correspondences between the query image and sparse SfM reconstruction of the scene, typically with hand-crafted local features like SIFT~\cite{lowe2004distinctive}. The pipeline scales up to large scenes using image retrieval~\cite{sivic2003video,arandjelovic2016netvlad, revaud2019learning, gordo2017end, ge2020self, yan2021image}, which restricts 2D-3D matching into visible parts of the query image. Recently, many traditional hand-crafted steps have been substituted by learning-based counterparts~\cite{detone2018superpoint, dusmanu2019d2, tian2019sosnet, sarlin2020superglue, sun2021loftr}. 
    HLoc~\cite{sarlin2019coarse,sarlin2020superglue} summarizes and provides a complete and leading toolbox for structure-based visual localization. However, during the retrieval phase, HLoc generally fails when the scene exhibits large visual variations, due to the limited information contained in global features. During the 2D-3D matching phase, HLoc is slow because it depends on multiple 2D-2D image matchings as the proxy, and it may still output a lot of outliers over notable appearance variations (weather, seasonal and illumination).
    For the retrieval and matching problem, we believe that leveraging other modality sensors could bring benefits. As for the time-consumption issue, inspired by recent 6-DoF pose estimation works~\cite{sun2022onepose,he2022onepose}, we aim to directly match 3D sub-map and 2D query image in one-shot with self- and cross-attention modules.
    
    \paragraph{Multi-sensor Visual Localization.}
    As GPS roughly provides absolute 3-DoF locations in outdoor environments, some methods employ the GPS signals as an extra constraint in optimization to improve the global accuracy of VIO~\cite{yu2019gps, schreiber2016vehicle, kim2014high, qin2019general} and visual SLAM~\cite{shi2012gps,chen2018integration,congram2021relatively}, while others treat GPS as a prior to simplify the image retrieval task for visual localization~\cite{li2018integration, vishal2015accurate, vysotska2015efficient}. 
    Gravity direction measured from IMU sensors is a commonly used prior for pose estimation considering its high accuracy~\cite{kukelova2010closed}. Previous researches usually extract the knowledge of known gravity direction to enhance the PnP solver by bootstrapping minimal solutions~\cite{kukelova2010closed,sweeney2015efficient,svarm2016city,albl2016rolling} or incorporating extra regularizers~\cite{sweeney2014gdls,fragoso2020gdls}.
    However, to our knowledge, no previous work has considered multiple sensors jointly. In fact, modern cell phones and other smart devices have been equipped with a large variety of sensors, including gyroscopes and accelerometers, compass, GPS, Wifi, Bluetooth and so on. A localization algorithm to fully exploit multiple sensors is desired. 
    
    \paragraph{Datasets.}
    The majority of existing datasets for visual localization do not provide~\cite{sattler2018benchmarking,zhang2021reference,jin2021image,kendall2015posenet} or just provide limited types of sensor data~\cite{chen2011city, shotton2013scene, taira2018inloc, sun2017dataset, carlevaris2016university, cortes2018advio, revaud2019learning, schops2017multi}. The closest to our work is the very recently released benchmark LaMAR~\cite{sarlin2022lamar} for augmented reality. However, LaMAR uses expensive and professional LiDAR devices to reconstruct the reference map and establish the ground-truth. Besides, LaMAR only provides radio signals from Wifi and Bluetooth, which are not always available in outdoor environments, and do not provide GPS and compass data. 
    
    To label the query GT, many previous works rely on offline SfM systems~\cite{schonberger2016structure} upon unstructured photo collections~\cite{sattler2018benchmarking, jin2021image, kendall2015posenet}. However, when it comes to the circumstance where a huge appearance change exists between query and reference images, merely adopting SfM algorithms may lead to limited registration accuracy. Some other datasets propose to use fiducial markers~\cite{ceriani2009rawseeds, carlevaris2016university, jeong2019complex}, depth information (e.g., RGB-D camera~\cite{shotton2013scene} or LiDAR~\cite{carlevaris2016university, taira2018inloc}) as additional pose constraints, or even introduce manual annotations~\cite{taira2018inloc, cortes2018advio, schops2017multi}. The requirement of external equipment and manual labeling greatly increase the capture cost and thus limit the dataset scalability. Recently, LSFB~\cite{liu2020lsfb} and LaMAR~\cite{sarlin2022lamar} employ an automatic and low-cost framework for AR devices' GT generation. Although we similarly fuse diverse sensor sources to automatically generate GT, we devise a solution for hard cases where query images exhibit dramatic appearance differences compared with the reference map. Our approach builds an auxiliary 3D map to simplify feature matching and introduces more constraints such as RTK positions and gravity directions.  
    
    A detailed comparison of representative outdoor datasets is given in Table~\ref{tab:dataset}.
    
    

     \section{Visual Localization with Mobile Sensors}
\begin{figure*}[t]
      \centering
      \includegraphics[width=0.95\linewidth]{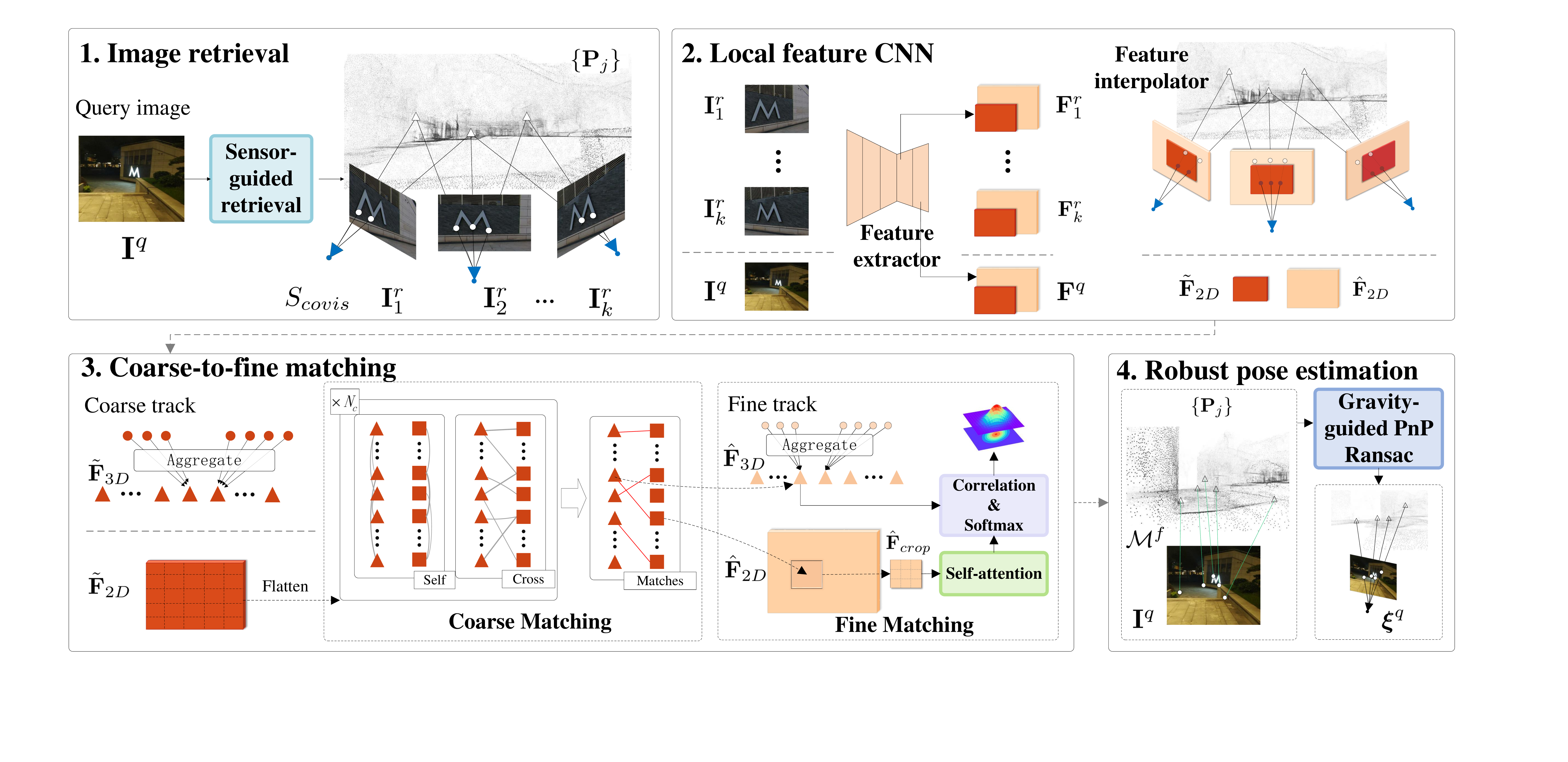}
    
      \caption{\textbf{Overview of the proposed method.} 1. For each query image $\mathbf{I}^q$, our sensor-guided retrieval algorithm first finds covisible references $S_{covis}$. 2. A local feature CNN extracts coarse-level and fine-level feature maps $\tilde{\mathbf{F}}_{2D}$ and $\hat{\mathbf{F}}_{2D}$. 3. 2D descriptors are aggregated to 3D descriptors for the feature tracks, followed by self- and cross-attention modules to build coarse correspondences and a fine matching module to refine these correspondences to sub-pixel positions. 4. The camera pose $\boldsymbol{\mathcal{\xi}}^q$ is estimated via gravity-guided PnP RANSAC.}
      \label{fig:main}
    \end{figure*}

    An overview of the proposed method is exhibited in \cref{fig:main}. Given the reference map consisting of images with known camera poses $\{\mathbf{I}^r,~\boldsymbol{\mathcal{\xi}}^r \}$ and point clouds $\{ \mathbf{P}_j \}$, our objective is to estimate the 6-DoF pose $\boldsymbol{\mathcal{\xi}}^q$ for the query image $\mathbf{I}^q$. To achieve this, we present a novel three-stage pipeline that first leverages GPS and compass to narrow the search space for image retrieval (Section~\ref{subsec:guided retrieval}). Then we build 2D-3D correspondences in a coarse-to-fine manner for query image $\mathbf{I}^q$ (Section~\ref{subsec:direct match}), and finally solve the camera pose $\boldsymbol{\mathcal{\xi}}^q$ by a gravity-guided PnP RANSAC (Section~\ref{subsec:guided pnp}).
    
    \subsection{Sensor-guided Image Retrieval}
    \label{subsec:guided retrieval}
    Given a query image $\mathbf{I}^q$, the retrieval stage aims to find a set of covisible images $S_{covis}= \{ \mathbf{I}^r_1,\mathbf{I}^r_2,...,\mathbf{I}^r_k \}$ among the reference image collection $\{ \mathbf{I}^r \}$, where $k$ represents the number of retrieved items. A commonly-used method is to first use an embedding function $\boldsymbol{f}(\cdot)$ to map images $\{ \mathbf{I}^q, \mathbf{I}^r \}$ into a compact feature space and then search for nearest neighbors of $\mathbf{I}^q$ using a distance metric $\boldsymbol{d}(\boldsymbol{f}(\mathbf{I}^q),\boldsymbol{f}(\mathbf{I}^r))$.

\begin{figure}[t]
       \centering
       \includegraphics[width=0.85\linewidth]{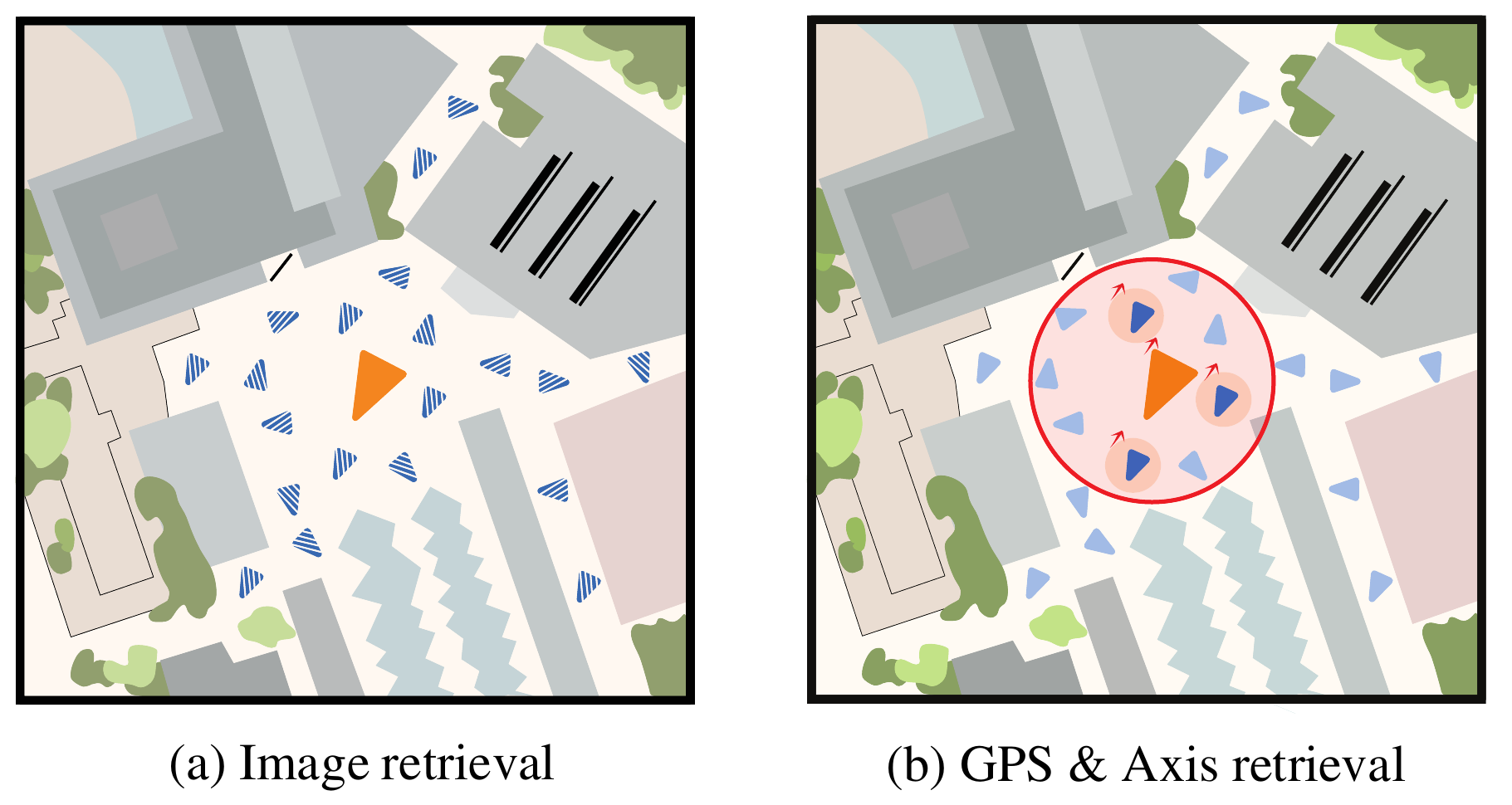}
       \caption{\textbf{Using the GPS and compass for image retrieval.} (a) Query views (orange triangle) against reference views (blue triangles). (b) Reducing search space by considering the GPS distance and axis direction.}
       \label{fig:retrieval}
    \end{figure}

    
    However, this method is completely dependent on the expressiveness of global features. When images are acquired from a site with significant visual changes, e.g., the \textit{SensLoc} dataset, it may fail. To address this issue, we propose to use the metadata from the mobile sensors as priors to filter incorrect candidates beforehand, as shown in \cref{fig:retrieval}. We denote the prior pose of the query image as $_s{\boldsymbol{\mathcal{\xi}}}^q$, which comprises translation from the built-in GPS signal and rotation from integrating the gravity and compass data. For each $\mathbf{I}^q$, we narrow the retrieval candidate set $^q\mathcal{I}$ as follows:
    \begin{equation}
    \begin{split}
    ^q\mathcal{I} = & \{ \mathbf{I}^r_i \mid \forall ~  \| \boldsymbol{\mathcal{\xi}}^r_i(\boldsymbol{t}_{xy}) - _s\boldsymbol{\mathcal{\xi}}^q(\boldsymbol{t}_{xy}) \| \leq \tau_t, \\
    & \arccos (\boldsymbol{\mathcal{\xi}}^r_i(\boldsymbol{p}) \cdot _s\boldsymbol{\mathcal{\xi}}^q(\boldsymbol{p})) \leq \tau_o \},
    \end{split}
    \label{equ:distance}
    \end{equation}
    where $\tau_t$ and $\tau_o$ are translation and orientation thresholds, respectively. We notice that the altitude value of GPS is unstable, so we only use the x-y coordinate $\boldsymbol{t}_{xy}$ in reference pose $\boldsymbol{\mathcal{\xi}}^r_i$ and query prior pose $_s\boldsymbol{\mathcal{\xi}}^q$ to constrain the search space. Besides, we compute the angle between principal axes of query pose prior ($_s\boldsymbol{\mathcal{\xi}}^q(\boldsymbol{p})$) and reference images ($\boldsymbol{\mathcal{\xi}}^r_i(\boldsymbol{p})$), to ensure that camera directions are similar. The camera’s principal axis is defined as a line perpendicular to the image plane that extends from the camera center. We use principal axes instead of calculating the camera orientations because images rotated around their principal axis still observe the same scene. Based on the filtering step, we can efficiently and accurately determine the k-nearest neighbors $S_{covis}$ according to $\boldsymbol{d}(\boldsymbol{f}(\mathbf{I}^q),\boldsymbol{f}(\mathbf{I}^p))$, where $\mathbf{I}^p \in ~^q\mathcal{I}$. The local point cloud $\{ \mathbf{P}_j \}$ observed by images in $S_{covis}$ are used for the latter feature matching step.

    \subsection{Direct 2D-3D Matching}
    \label{subsec:direct match}
    For a query $\mathbf{I}^q$, assuming that we have obtained its covisible neighbors $S_{covis}$ and relevant point cloud $\{ \mathbf{P}_j \}$, we focus on establishing 2D-3D matches between sub-pixels in $\mathbf{I}^q$ and points in $\{ \mathbf{P}_j \}$.
    
    \paragraph{Preprocess.}
    First, we hierarchically extract coarse- and fine-level dense feature maps $\{ \mathbf{F}^q, \{ \mathbf{F}^r_1,\mathbf{F}^r_2,...,\mathbf{F}^r_k \}\}$ from images $\{\mathbf{I}^q, \{ \mathbf{I}^r_1,\mathbf{I}^r_2,...,\mathbf{I}^r_k \}\}$. For clarity, we use $\tilde{\cdot}$ and $\hat{\cdot}$ to denote the coarse-level and fine-level features, so that extracted multi-level features can be denoted as $\tilde{\mathbf{F}}_{2D} \in \mathbb{R}^{\frac{H}{8} \times \frac{W}{8} \times C_c}$ for the coarse level and $\hat{\mathbf{F}}_{2D} \in \mathbb{R}^{\frac{H}{2} \times \frac{W}{2} \times C_f}$ for the fine level, where $H, W$ indicate image height and width, and $C_c, C_f$ represent the dimensions of the coarse and fine features, respectively. Next, every point $\mathbf{P}_j$ is reprojected into the feature maps in $S_{covis}$ and the corresponding feature descriptors are extracted using bilinear interpolation. This step yields a feature track $\mathcal{T}_j = \{ (\tilde{\mathbf{F}}_{2D}(k), \hat{\mathbf{F}}_{2D}(k)) | k=1...N_{j_k}\}$ for each 3D point $\mathbf{P}_j$, where $N_{j_k}$ is the track length. Finally, an average pooling operation is used for each track $\mathcal{T}_j$ to form a 3D descriptor $(\tilde{\mathbf{F}}_{3D}(j), \hat{\mathbf{F}}_{3D}(j))$. 
    
    \paragraph{Coarse-to-fine matching.}
    Based on the 2D and 3D features $\{ \tilde{\mathbf{F}}_{2D}, \hat{\mathbf{F}}_{2D} \}$ and $\{ \tilde{\mathbf{F}}_{3D}, \hat{\mathbf{F}}_{3D} \}$, we initially determine whether a 3D point in sub-map $\{ \mathbf{P}_j \}$ is observable by the query $\mathbf{I}^q$ and search for a rough location $\tilde{\mathbf{u}}^i$ in the coarse level. Afterward, the coarse correspondence is refined to a sub-pixel position $\hat{\mathbf{u}}^i$ by a fine-matching module. 
    
    Similar to~\cite{sun2021loftr, he2022onepose}, we first use positional encoding to augment the features $\{ \tilde{\mathbf{F}}_{2D}, \tilde{\mathbf{F}}_{3D} \}$, resulting in $\{ \tilde{\mathbf{F}}_{2D}^{pe}, \tilde{\mathbf{F}}_{3D}^{pe} \}$. This process involves 2D pixel extension following~\cite{Carion2020EndtoEndOD} and 3D coordinate encoding with Multilayer Perceptron (MLP)~\cite{pinkus1999approximation}. We flatten the 2D coarse map $\tilde{\mathbf{F}}_{2D}^{pe}$ and apply linear self-attentions and cross-attentions~\cite{Katharopoulos2020TransformersAR} upon $\{ \tilde{\mathbf{F}}_{2D}^{pe}, \tilde{\mathbf{F}}_{3D}^{pe} \}$ $N_c$ times to obtain transformed features $\{ \tilde{\mathbf{F}}_{2D}^t, \tilde{\mathbf{F}}_{3D}^t \}$. Subsequently, a probability matrix $\mathcal{P}^{c}$ is calculated by a dual-softmax operation on the correlation volume $\mathrm{C}$, which is derived from  the inner product of 2D and 3D features $\{ \tilde{\mathbf{F}}_{2D}^t, \tilde{\mathbf{F}}_{3D}^t \}$: 
    \begin{equation}
    \begin{split}
    \label{eq:dual-softmax}
        \mathcal{P}^{c}(i, j) = \operatorname{softmax}\left(\mathrm{C}\left(i, \cdot \right)\right)_j \cdot \operatorname{softmax}\left(\mathrm{C}\left(\cdot, j\right)\right)_i, \\
        \text{where}~\mathrm{C}\left(i, j\right) = \frac{1}{\tau} \cdot \langle\tilde{\mathbf{F}}_{2D}^t(i),~\tilde{\mathbf{F}}_{3D}^t(j)\rangle.
    \end{split}
    \end{equation}
    $\langle \cdot, \cdot \rangle$ indicates the inner product operation, $\tau$ represents the temperature parameter of $\operatorname{softmax}$, and $i$, $j$ are the indices of a pixel in the flattened coarse map and a 3D point, respectively. The coarse 2D-3D correspondences, denoted as ${\mathcal{M}}^c$, are constructed from $\mathcal{P}^{c}$ by selecting correspondences that meet a confidence threshold $\theta$ and satisfy the mutual nearest neighbor ($\operatorname{MNN}$) criteria:
    \begin{equation}
    \resizebox{0.9\linewidth}{!}{
        $
        \mathcal{M}^{c} = \{\left(i,j\right) \mid \forall\left(i,j\right) \in \operatorname{MNN}\left(\mathcal{P}^{c}\right),\ \mathcal{P}^{c}\left(i,j\right) \geq \theta\},
        \label{eq:coarse-matches}
        $
    }
    \end{equation}
    
    For each coarse correspondence $(\tilde{\mathbf{u}}^i,\mathbf{P}^j)$ identified by ${\mathcal{M}}^c$, a local window $\hat{\mathbf{F}}_{crop}$ of size $ w \times w $ is cropped around $\tilde{\mathbf{u}}^i$ in the fine feature $\hat{\mathbf{F}}_{2D}$. A similar self-attention and cross-attention layer is then applied to transform the cropped feature map $\hat{\mathbf{F}}_{crop}$ and its corresponding 3D fine feature $\hat{\mathbf{F}}_{3D}(j)$ $N_f$ times. Finally, we correlate the transformed 3D fine feature $\hat{\mathbf{F}}_{3D}^t(j)$ with all elements in the transformed crop feature $\hat{\mathbf{F}}_{crop}^t$ to produce a heatmap representing the matching probability in the local region. By estimating the 2D expectation over the probability distribution, we gain the fine correspondence position $\hat{\mathbf{u}}^i$ with respect to $\mathbf{P}^j$. 
    
    Following ~\cite{sun2021loftr, he2022onepose}, we supervise the network by a focal loss~\cite{linFocalLossDense2018} for coarse matching and variance-weighted $\ell_2$ loss for fine refinement. More details are provided in the supplementary material.


    \subsection{Gravity-guided PnP RANSAC}
    \label{subsec:guided pnp}
    After establishing the 2D-3D fine matches ${\mathcal{M}}^f$ between the query image $\mathbf{I}^q$ and the local point cloud $\mathbf{P}^j$, previous works commonly estimate the pose $\boldsymbol{\mathcal{\xi}}^q$ with the Perspective-n-Point~(PnP)~\cite{Lepetit2008EPnPAA} technique in a RANSAC~\cite{fischler1981random} framework. 
    
    However, such methods may have trouble with scenes containing large illumination changes as they highly rely on the performance of local feature matching. To this end, we propose to insert a simple yet effective verification module into a LO-RANSAC~\cite{chum2003locally} scheme to ensure the correctness of the gravity direction. In detail, for each pose hypothesis $\boldsymbol{\mathcal{\xi}}_{hyp}$ estimated from a random sample of three 2D-3D correspondences, we compute the difference of gravity directions $d_\epsilon$ between sensor pose $_s\boldsymbol{\mathcal{\xi}}$ and hypothesis pose $\boldsymbol{\mathcal{\xi}}_{hyp}$, as 
    \begin{equation}
    \resizebox{0.55\linewidth}{!}{
        $
        d_\epsilon = \arccos (_s\boldsymbol{\mathcal{\xi}}(\boldsymbol{g}) \cdot \boldsymbol{\mathcal{\xi}}_{hyp}(\boldsymbol{g})).
        \label{eq:gravity pnp}
        $
    }
    \end{equation}
    The notations $_s\boldsymbol{\mathcal{\xi}}(\boldsymbol{g})$ and $\boldsymbol{\mathcal{\xi}}_{hyp}(\boldsymbol{g})$ refer to the gravity directions of sensor and hypothesis poses. During RANSAC iterations, an earlier termination is permitted if $d_\epsilon \geq \tau_\epsilon$, where $\tau_\epsilon$ is a predefined maximum gravity error.
    
    According to this strict criterion, we not only suppress potential false hypotheses and return high-quality results, but also could handle overload situations when there are too many tentative 2D-3D correspondences. This is achieved by filtering out a large fraction of incorrect poses with $d_\epsilon \geq \tau_\epsilon$ in advance.
        
    \subsection{Implementation details}
    \label{subsec:implement details}
    
    ResNet-18~\cite{He2016DeepRL} is used as the feature extractor and we set $N_c$ to 4 and $N_f$ to 1 for the transformer modules. The temperature value $\tau$ is 0.1. The crop window size $w$ is equal to $5$. As for training, we select the same sequences on MegaDepth~\cite{li2018megadepth} with \cite{dusmanu2019d2}, but retriangulate the 3D structure with Superpoint~\cite{detone2018superpoint} and Superglue~\cite{sarlin2020superglue} to increase the point density. We randomly sample or pad 10 covisible images $S_{covis}$ and extract a visible local sub-map $\{ \mathbf{P}_j \}$ with $4000$ points for batch fetching. The backbone of our model is fixed with the LoFTR-DS~\cite{sun2021loftr} outdoor model, while the other parts are randomly initialized and trainable. The model is trained using AdamW with an initial learning rate of $8\times10^{-3}$ and a batch size of $24$ on $4$ NVIDIA GTX 3090 GPUs. During testing, we set the confidence threshold $\theta$ to $0.05$ in the daytime and $0.005$ in the night-time, as SensLoc has the ability to handle tentative matches brought by such a low threshold. The translation and orientation thresholds $\tau_t$ and $\tau_o$ in Eq.~\ref{equ:distance} are $20$ meters and $60$ degrees for sensor-guided retrieval. The gravity error threshold $\tau_\epsilon$ is set to $2$ degrees for gravity-guided PnP RANSAC. All testing experiments are operated on an NVIDIA TITAN RTX GPU.

    \section{Dataset}

    \begin{figure*}[t]
      \centering
      \includegraphics[width=0.98\linewidth]{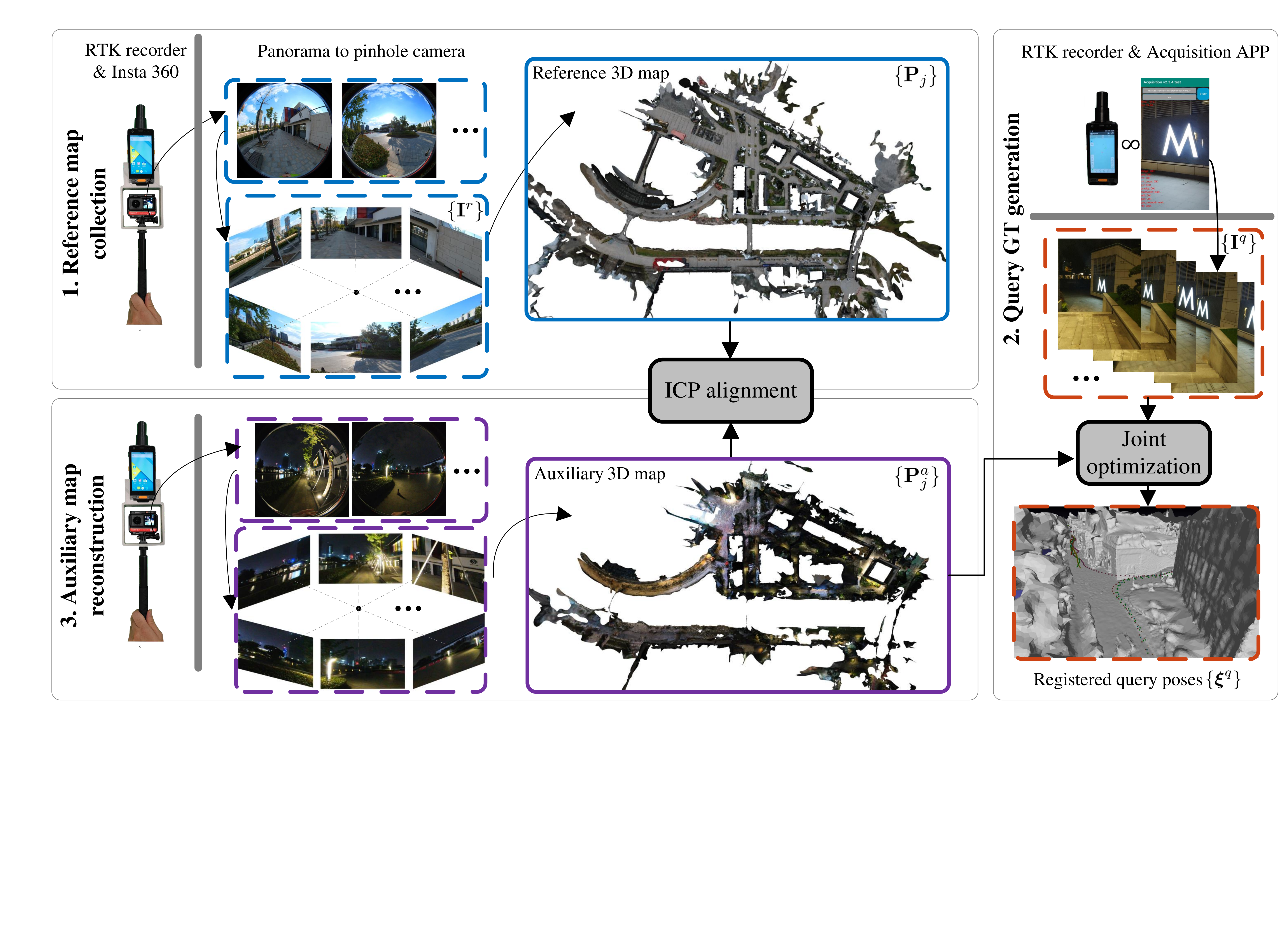}
    
      \caption{\textbf{Overview of dataset collection.} The dataset construction can be divided into three parts: 1. Building a large-scale reference map with Insta 360 and RTK. 2. Collecting query images by a mobile phone bounded with an RTK recorder. 3. At the same time with query image collection, building an auxiliary map with Insta 360 and RTK to facilitate query-reference matching. The GT poses of query images are generated by first aligning the auxiliary map to the reference map by ICP and then registering query images to the auxiliary map by joint optimization based on multi-sensor information.}
      \label{fig:dataset2}
    \end{figure*}

    The released dataset \textit{SensLoc} includes a large city park (approximately $31,250 m^2$), containing vegetation, rivers, buildings and walkways. As a publicly accessible area, unavoidably, \textit{SensLoc} experiences various scene changes all the time, e.g., moving people, vehicles or even new infrastructure constructions. The query images were captured with different illuminations, seasons and weather, compared to the reference images. A visualization of the dataset can be found in \cref{fig:dataset1}. The supplementary material provides statistics on the number of query images for each condition.
 
    \subsection{Reference map collection}
    We created a digital map upon \textit{SensLoc} environment with a panoramic camera, Insta360 ONE\footnote{www.insta360.com/product/insta360-oners}, due to the efficiency of 3D reconstruction from panoramic imaging over monocular acquisition. Specifically, given reference images $\{ \mathbf{I}^r \}$ captured by Insta360, we applied SfM techniques to reconstruct sparse point cloud $\{\mathbf{P}_j \in \mathbb{R}^3 \}$ and feature track $\{\mathcal{T}_j\}$, where $j$ is the point index. After converting $7,958$ database images from panorama to pinhole, \textit{SensLoc} dataset results in $47,748$ reference images and $0.31M$ 3D points triangulated from $6.20M$ local features. In order to recover the scale of the reconstruction and align it with the real geographic world, we pre-calibrated and equipped the Insta360 ONE with a RTK recorder, T38P-E, to document the geographical locations. The pipeline of the reference map reconstruction is illuminated in the top area of \cref{fig:dataset2}.

    \subsection{Query image collection}
    For query image collection, we freely walked through the \textit{SensLoc} environment several times after half a year, and captured videos using Huawei P40pro and Xiaomi Mix3 smartphones with a developed Android-powered Acquisition Application (APP). Since the SensLoc algorithm focuses on image localization, we sampled all sequences into single frames and totally generated $1,576$ day-time and $362$ night-time queries. Note that without downsampling, our dataset can also support sequence localization tasks. The Acquisition APP records raw data that measure motion, orientation, and various environmental conditions from the phones' built-in sensors. It includes but not limited to accelerometer, gyroscope, gravity, compass, Bluetooth, WiFi and GPS measurements. Please refer to the supplementary material for 
    a detailed introduction. Besides, we employed a handheld RTK device attached to the phone rigidly to record shooting locations, which could provide GT locations at centimeter-level accuracy~\cite{tradacete2018positioning}. All sensors have been hardware synchronized and carefully calibrated. 

    

    \paragraph{Auxiliary map reconstruction.} 
    We found that it was quite difficult to directly generate pseudo-GT poses against the pre-built reference map $\{ \mathbf{P}_j \}$ due to large appearance changes between query and reference images. 
    To solve this problem, we propose to reconstruct an auxiliary 3D map $\{ \mathbf{P}_j^a \}$ during the query time. This process is detailed in the median part of \cref{fig:dataset2}. Specifically, we first replicate the reference map construction process by utilizing the Insta360 and RTK recorder to capture and build a 3D auxiliary model. The auxiliary model $\{ \mathbf{P}_j^a \}$ is then aligned with the base reference model $\{\mathbf{P}_j \}$. In detail, both $P_j$ and $P^a_j$ are initially aligned with the real geographic world by RTK with an error less than 0.5m. Fine alignment is achieved using the ICP~\cite{rusinkiewicz2001efficient}, which is not likely to get stuck in local minima due to the good initialization. The RMSE of all inlier 3D-3D matches is around 1.5cm and the alignment quality can be visually inspected in supplementary materials. Based on the alignment of two maps, pseudo-GT generation in $\{ \mathbf{P}_j \}$ is equal to recovering accurate poses in $\{ \mathbf{P}_j^a \}$, where registering query images to a 3D map collected at the same time is much easier.


    \paragraph{Query GT generation.}
    We apply a joint optimization algorithm to find the pseudo-GT pose parameters $\{ \boldsymbol{\mathcal{\xi}}_i^q \}$ that meet the requirements: 1) the camera self-localization, 2) motion constraints from IMU measurements, and 3) sensor prior constraints regarding translation and orientation.
    The self-localization recovers camera poses $\{ \boldsymbol{\mathcal{\xi}}_i^q \}$ by finding 2D-3D correspondences between point cloud $\{ \mathbf{P}_j^a \}$ and query image $\{ \mathbf{I}^q \}$. To impose temporal constraints, VIO is employed as another constraint to refine the localization $\{ \boldsymbol{\mathcal{\xi}}_i^q \}$. To further increase the pose accuracy, we use the measured RTK positions and gravity directions as regularization. 
    The entire GT generation is formulated as the following joint optimization~\cite{triggs1999bundle} function:
    \begin{equation}
    \begin{split}
    & \boldsymbol{\mathcal{\xi}}_i^{q*} = \underset{\boldsymbol{\mathcal{\xi}}_i^q, \mathbf{X}_j}{\argmin} \\
    & w_{sl} \underset{i}{\sum} \underset{j}{\sum} \| \mathbf{p}_{ij}^q - \boldsymbol{\pi} \left( \boldsymbol{\mathcal{\xi}}_i^q, \mathbf{P}_j \right) \| _{\sigma^2_{sl}} + \\
    & w_{vo}  \underset{i}{\sum} \underset{j}{\sum} \| \mathbf{x}_{ij}^q - \boldsymbol{\pi} \left( \boldsymbol{\mathcal{\xi}}_i^q, \mathbf{X}_j \right) \| _{\sigma^2_{vo}} + \\
    & w_{io} \underset{i}{\sum} \| \boldsymbol{h} \left( \boldsymbol{\mathcal{\xi}}_i^q, \boldsymbol{\mathcal{\xi}}_{i+1}^q, \boldsymbol{v}_i, \boldsymbol{b}_a, \boldsymbol{b}_g \right) \| _{\sigma^2_{io}} + \\
    & w_t \underset{i}{\sum} \| \boldsymbol{\dot{t}}_i^{q} - \boldsymbol{t}_i^q \| _{\sigma^2_{t}} + w_g \underset{i}{\sum} (\arccos (\boldsymbol{\dot{g}}_i^q \cdot \boldsymbol{g}_i^q)) _{\sigma^2_{g}},
    \end{split}
    \end{equation}
    where $ \boldsymbol{\pi}(\cdot) $ represents a pinhole projection function. The sets $\{ (\mathbf{P}_j, \mathbf{p}_{ij}^q) \}$ and $\{ (\mathbf{X}_j, \mathbf{x}_{ij}^q) \}$ indicate 2D-3D matches in self-localization and visual odometry respectively. The energy function $\boldsymbol{h}(\cdot)$~\cite{forster2015imu} is used to measure the difference between consecutive poses $\{ \boldsymbol{\mathcal{\xi}}_i^q, \boldsymbol{\mathcal{\xi}}_{i+1}^q \}$ and the integration of IMU measurements $\{ \boldsymbol{b}_a,\boldsymbol{b}_g \}$ and velocity $\boldsymbol{v}_i$. We employ the x-y coordinate value $\{ \boldsymbol{\dot{t}}^{q}_{i} \}$ from RTK to direct the movement of camera trajectory $\{ \boldsymbol{t}^q_i \}$, where  $\boldsymbol{t}^q_i = \boldsymbol{\mathcal{\xi}}_i^q(\boldsymbol{t}_{xy})$. Furthermore, we impose a misalignment penalty on query gravity direction $\{ \boldsymbol{g}_i^q \}$ based on reliable gravity sensor output $\{ \boldsymbol{\dot{g}}_i^q \}$, where $ \boldsymbol{g}_i^q = \boldsymbol{\mathcal{\xi}}_i^q(\boldsymbol{g}) $. The term $||\mathbf{e}||_\sigma^2 = \mathbf{e}^T\mathbf{e} / \sigma^2$ represents the squared norm of the error vector. Further details can be found in the supplementary material. 
    
    The major errors in our generated GT may include: the error of the RTK recorder (1$cm$ to 3$cm$), calibration error between the RTK recorder and mobile phone (approximately 2$cm$), and misalignment between the reference and auxiliary maps (approximately 1.5$cm$), which make our pseudo-GT accuracy lie in the decimeter level.

    \section{Experiments}
    \begin{table} [t]
    	\centering
    	\resizebox{0.47\textwidth}{!}{%
    		\setlength\tabcolsep{2pt}
    		\begin{tabular}{cc|ccccccc}
    			\Xhline{3\arrayrulewidth}
    			Global feature & Prior & R@1 & R@5 & P@5 & R@10 & P@10 & R@20 & P@20 \\
    			\midrule
    			- & Axis+GPS & 34.73 & 77.29 & 34.20 & 82.04 & 34.38 & 87.20 & 33.91  \\
    			\midrule
    			\multirow{4}*{AP-GeM~\cite{gordo2017end}} & - & 51.87 & 67.75 & 44.46 & 74.45 & 39.60 & 79.71 & 32.58 \\
    			& Axis & 53.10 & 71.67 & 47.63 & 79.31 & 43.16 & 85.35 & 36.80 \\ 
    			& GPS & 61.56 & 82.40 & 54.91 & 89.27 & 50.03 & 93.81 & 43.12 \\ 
    			& Axis+GPS & \textbf{64.96} & \textbf{88.13} & \textbf{60.27} & \textbf{93.19} & \textbf{55.92} &  \textbf{96.70} & \textbf{49.59} \\ 
    			\midrule
    			\multirow{4}*{NetVLAD~\cite{arandjelovic2016netvlad}} & - & 59.80 & 73.87 & 53.87 & 77.80 & 48.10 & 82.20 & 38.97 \\
    			& Axis & 63.57 & 80.90 & 58.26 & 85.81 & 52.77 & 89.83 & 43.75 \\ 
    			& GPS & 70.85 & 90.30 & 66.21 & 94.69 & 61.08 & 97.06 & 52.39 \\
    			& Axis+GPS & \textbf{74.51} & \textbf{93.29} & \textbf{70.33} & \textbf{96.59} & \textbf{65.44} & \textbf{98.91} & \textbf{57.27} \\ 		
    			\midrule
    			\multirow{4}*{OpenIBL~\cite{ge2020self}} & - & 68.42 & 82.49 & 63.44 & 86.03 & 57.58 & 89.19 & 47.55 \\
    			& Axis & 68.57 & 83.75 & 63.81 & 87.82 & 58.02 & 91.07 & 48.92 \\ 
    			& GPS & 75.12 & 91.59 & 70.99 & 95.77 & 66.16 & 97.88 & 57.09 \\ 
    			& Axis+GPS & \textbf{77.50} & \textbf{92.93} & \textbf{73.49} & \textbf{96.65} & \textbf{69.22} & \textbf{98.56} & \textbf{60.73} \\ 
    			\Xhline{3\arrayrulewidth}
    		\end{tabular}%
    	}
    	\caption{\textbf{Image retrieval results.} We report the top-$k$ recall and precision when $k=1,5,10,20$ using different image features and sensor priors. A retrieval result is deemed correct if it falls within a distance of $\tau_t=10m$ from the query view and the orientation between them is less than $\tau_o = 30^{\circ}$.}
    	\label{tab:retrieval}
    \end{table}

    \subsection{Image Retrieval}
    \label{subsec:retrieval}
    
    \paragraph{Baselines and metrics.} 
    We evaluate global descriptors computed by AP-GeM~\cite{gordo2017end}, NetVLAD~\cite{arandjelovic2016netvlad} and OpenIBL~\cite{ge2020self}, which are representative of the field~\cite{pion2020benchmarking}. In order to learn the effect of the provided priors (i.e., axis direction and GPS xy position), we carry out controlled experiments on the basis of distinct global features. 
    For evaluation, a retrieval result is deemed correct if it satisfies two requirements: 1) it is within $\tau_t=10$ meters from the query position, and 2) it shares a similar perspective with the query pose, where $\tau_o \textless 30^{\circ}$. We report recall and precision metrics on the behalf of the top $k$ retrieved items, and $k$ ranges from $1$ to $20$. 
    
    \paragraph{Results.}
    The retrieval results are presented in Table~\ref{tab:retrieval}. Even though sensor prior itself cannot output accurate results, it always improves the retrieval performance over vanilla feature-only methods, regardless of the top $k$ threshold. Combining the global feature OpenIBL with Axis (orientation) and GPS (position) priors leads the benchmark, which is adopted to find the covisible reference set for the following localization methods.

    \begin{table*}[t]
    \centering
    \resizebox{0.8\textwidth}{!}{
    \setlength\tabcolsep{2pt} 
    	\begin{tabular}{l||ccc|ccc||cc} 
    		\Xhline{3\arrayrulewidth}
    		\multirow{2}{*}{} & \multicolumn{3}{c|}{Day} & \multicolumn{3}{c||}{Night} & \multirow{2}{*}{Time~(Net)} & \multirow{2}{*}{Time~(PnP)} \\ 
    		\cline{2-7} & $(25cm, 2^{\circ})$ & $(50cm, 5^{\circ})$ &  $(1m, 10^{\circ})$ & $(25cm, 2^{\circ})$ & $(50cm, 5^{\circ})$ &  $(1m, 10^{\circ})$ & \\ 
    		\hline
    		\hline
    		HLoc~\textit{(SIFT + NN)} & 62.56 & 74.56 & 80.52 & 22.65 & 29.28 & 35.08 & 706 & 18
                                                                         \\
    		HLoc~\textit{(SPP + NN)} & 73.73 & 83.19 & 86.99 & 34.81 & 39.23 & 41.71 & 1361 & 18
                                                                         \\
    		HLoc~\textit{(D2Net + NN)} & 16.31 & 25.06 & 32.74 & 1.10 & 2.76 & 3.87 & 1987 & 10
                                                                         \\
    		HLoc~\textit{(SPP + SPG)} & \textbf{86.04} & \textbf{93.34} & \textbf{95.18} & 70.99 & 77.90 & 	80.94 & 2419 & 92
                                                                         \\
            HLoc~\textit{(LoFTR${}^{*}$)} & 78.81 & 89.53 & 92.70 & 72.65 & 87.85 & 91.16 & 2880 & 170                                                            \\
    		\hline
    		Pixloc~\textit{(Retrieval initial)} & 31.85 & 37.75 & 43.78 & 20.99 & 28.18 & 35.08 & \multicolumn{2}{c}{1237~\textit{(Direct alignment)}}
                                                              \\
    		\hline
    		SensLoc                   & 85.09 & 93.02 & 94.99 & 80.94 & 92.27 & 93.65 & \textbf{66} & 33
                                                               \\
    		SensLoc + Gravity         & 85.28 & 93.02 & 95.05 & \textbf{81.49} & \textbf{92.27} & \textbf{94.48} & \textbf{66} & \textbf{8}
                                                               \\
    		\Xhline{3\arrayrulewidth}
    	\end{tabular}
    }
    \caption{\textbf{Visual localization results.} Our method is compared with HLoc~\cite{sarlin2019coarse} with different feature matchers and Pixloc~\cite{sarlin2021back}. The ``Net'' and ``PnP'' times refer to the running time in $ms$ of the feature matching and PnP RANSAC processes, respectively. Note that all the methods use the same multi-sensor retrieval method, i.e., OpenIBL+Axis+GPS in Table~\ref{tab:retrieval}. }
    \label{tab:localization}
    \end{table*}

    \subsection{Visual Localization}
    \label{subsec:localization}
    
    \paragraph{Baselines and metrics.}
    We compare our approach with the following baselines in two categories: 1) Feature matching pipelines HLoc~\cite{sarlin2019coarse}, using different keypoint descriptors (SIFT~\cite{lowe2004distinctive}, SuperPoint~\cite{detone2018superpoint} and D2-Net~\cite{dusmanu2019d2}), and matchers (Nearest Neighbour and learned SuperGlue~\cite{sarlin2020superglue}), as well as the detector-free matcher LoFTR~\cite{sun2021loftr}. 2) The end-to-end direct alignment approach Pixloc~\cite{sarlin2021back}.
    Note that LoFTR outputs semi-dense correspondence between reference images, which brings unacceptable memory overhead to BA optimization in our large-scale dataset \textit{SensLoc}. We thus implement an interval sampling upon raw tentative matches for sparse point cloud reconstruction (denoted as HLoc (\textit{LoFTR${}^{*}$})).
    We follow the standard localization evaluation procedure~\cite{toft2020long}, and report the localization recall at thresholds $(25cm, 2^{\circ})$, $(50cm, 5^{\circ})$, and $(1m, 10^{\circ})$. The testing time-cost is also compared in terms of network inference and PnP RANSAC, as it is an important indicator for the navigation performance of mobile devices in the real world.
 
    \paragraph{Results.}
    We report the results in Table~\ref{tab:localization}. The state-of-the-art method HLoc (SuperPoint+SuperGlue) obtains the best results during the daytime. It is reasonable because well-lit images have rich textures, benefiting the keypoint detection of SuperPoint. In the challenging condition of night, our approach outperforms other baselines by a large margin thanks to the keypoint-free design of 2D-3D matching. In addition, as we execute sparse-to-dense correspondence searching in one-shot, the network runs about 30x faster than the second competitor SuperPoint+SuperGlue. Furthermore, it is obvious that our proposed gravity validation module not only improves the localization accuracy, but also speeds up the PnP RANSAC process by 4 times.
    As we can see, even in severe long-term conditions, according to tightly coupling visual localization with multiple mobile sensors, we could obtain a satisfactory pose in real-time on a desktop computer.

    \subsection{Ablation Studies}
    \label{subsec:ablation studies}
    We conduct experiments to exploit the sensitivity of SensLoc to image retrieval, as shown in \cref{tab:ablation_study}. Increasing the retrieval accuracy using sensor-guided retrieval or ground-truth markedly increases the localization recall at $(25cm, 2^{\circ})/(50cm, 5^{\circ})/(1m, 10^{\circ})$. The ablation study demonstrates that the retrieval stage, which selects local point clouds in Section~\ref{subsec:guided retrieval}, plays a critical role in the pose estimation of our method. More ablation studies are provided in the supplementary material.

    \begin{table}[t]
	\centering
	\resizebox{0.48\textwidth}{!}{%
        \setlength\tabcolsep{4.0pt}
		\begin{tabular}{lcc}
			\Xhline{3\arrayrulewidth}
            \multirow{2}{*}{Retrieval} & Day & Night \\
            \cline{2-3} 
            &\multicolumn{2}{c}{$(25cm, 2^{\circ})/(50cm, 5^{\circ})/(1m, 10^{\circ})$} \\
            \hline
            w/o sensor-guided & 75.38 / 81.54 / 82.87 & 57.46 / 62.15 / 64.64  \\
            w sensor-guided & 85.28 / 93.02 / 95.05 & 81.49 / 92.27 / 94.48  \\ 
            ground-truth & 87.44 / 95.81 / 97.27 & 87.02 / 93.92 / 96.41  \\
            \Xhline{3\arrayrulewidth}
		\end{tabular}%
	}
    \addtolength{\textfloatsep}{-0.2in}
    \caption{\textbf{Ablation study.} We quantitatively evaluate the influence of image retrieval on SensLoc.}
    \label{tab:ablation_study}
    \end{table}

    

    
    \section{Conclusion}
\label{sec:conclusion}

This paper presents SensLoc, a new approach for long-term visual localization based on multi-modality sensors on mobile phones. To overcome the difficulties of image matching in temporally-varying outdoor environments, we propose to leverage additional mobile sensors, mainly GPS, compass and gravity sensors, to assist both image retrieval and pose estimation. We also introduce a new outdoor dataset that provides a variety of mobile sensor data and strong appearance changes between query and reference images. Our method significantly outperforms the state-of-the-art localization methods in terms of both accuracy and time-cost.

\paragraph{Acknowledgements.} 
The authors would like to acknowledge the support from the National Key Research and Development Program of China (No. 2020AAA0108901) and NSFC (No. 62171451, No. 62101576, No. 62071478).
    
    {\small
    \bibliographystyle{ieee_fullname}
    \bibliography{egbib}

\begin{thebibliography}{10}\itemsep=-1pt

\bibitem{albl2016rolling}
Cenek Albl, Zuzana Kukelova, and Tomas Pajdla.
\newblock Rolling shutter absolute pose problem with known vertical direction.
\newblock In {\em CVPR}, 2016.

\bibitem{arandjelovic2016netvlad}
Relja Arandjelovic, Petr Gronat, Akihiko Torii, Tomas Pajdla, and Josef Sivic.
\newblock Netvlad: Cnn architecture for weakly supervised place recognition.
\newblock In {\em CVPR}, 2016.

\bibitem{barath2019magsac}
Daniel Barath, Jiri Matas, and Jana Noskova.
\newblock {MAGSAC:} marginalizing sample consensus.
\newblock In {\em CVPR}, 2019.

\bibitem{brachmann2021visual}
Eric Brachmann and Carsten Rother.
\newblock Visual camera re-localization from {RGB} and {RGB-D} images using
  {DSAC}.
\newblock {\em T-PAMI}, 2021.

\bibitem{Carion2020EndtoEndOD}
Nicolas Carion, Francisco Massa, Gabriel Synnaeve, Nicolas Usunier, Alexander
  Kirillov, and Sergey Zagoruyko.
\newblock End-to-end object detection with transformers.
\newblock In {\em ECCV}, 2020.

\bibitem{carlevaris2016university}
Nicholas Carlevaris{-}Bianco, Arash~K. Ushani, and Ryan~M. Eustice.
\newblock University of michigan north campus long-term vision and lidar
  dataset.
\newblock {\em IJRR}, 2016.

\bibitem{castle2008video}
Robert Castle, Georg Klein, and David~W. Murray.
\newblock Video-rate localization in multiple maps for wearable augmented
  reality.
\newblock In {\em ISWC}, 2008.

\bibitem{ceriani2009rawseeds}
Simone Ceriani, Giulio Fontana, Alessandro Giusti, Daniele Marzorati, Matteo
  Matteucci, Davide Migliore, Davide Rizzi, Domenico~G Sorrenti, and Pierluigi
  Taddei.
\newblock Rawseeds ground truth collection systems for indoor self-localization
  and mapping.
\newblock {\em Auton. Robots}, 2009.

\bibitem{chen2011city}
David~M. Chen, Georges Baatz, Kevin K{\"{o}}ser, Sam~S. Tsai, Ramakrishna
  Vedantham, Timo Pylv{\"{a}}n{\"{a}}inen, Kimmo Roimela, Xin Chen, Jeff Bach,
  Marc Pollefeys, Bernd Girod, and Radek Grzeszczuk.
\newblock City-scale landmark identification on mobile devices.
\newblock In {\em CVPR}, 2011.

\bibitem{chen2018integration}
Xiao Chen, Weidong Hu, Lefeng Zhang, Zhiguang Shi, and Maisi Li.
\newblock Integration of low-cost {GNSS} and monocular cameras for simultaneous
  localization and mapping.
\newblock {\em Sensors}, 2018.

\bibitem{cheng2019cascaded}
Wentao Cheng, Weisi Lin, Kan Chen, and Xinfeng Zhang.
\newblock Cascaded parallel filtering for memory-efficient image-based
  localization.
\newblock In {\em ICCV}, 2019.

\bibitem{chum2008optimal}
Ond{\v{r}}ej Chum and Ji{\v{r}}{\'\i} Matas.
\newblock Optimal randomized {RANSAC}.
\newblock {\em T-PAMI}, 2008.

\bibitem{chum2003locally}
Ond{\v{r}}ej Chum, Ji{\v{r}}{\'\i} Matas, and Josef Kittler.
\newblock Locally optimized {RANSAC}.
\newblock In {\em DAGM}, 2003.

\bibitem{congram2021relatively}
Benjamin Congram and Timothy~D. Barfoot.
\newblock Relatively lazy: Indoor-outdoor navigation using vision and gnss.
\newblock In {\em CRV}, 2021.

\bibitem{cortes2018advio}
Santiago Cort{\'e}s, Arno Solin, Esa Rahtu, and Juho Kannala.
\newblock {ADVIO:} an authentic dataset for visual-inertial odometry.
\newblock In {\em ECCV}, 2018.

\bibitem{detone2018superpoint}
Daniel DeTone, Tomasz Malisiewicz, and Andrew Rabinovich.
\newblock Superpoint: Self-supervised interest point detection and description.
\newblock In {\em CVPRW}, 2018.

\bibitem{dusmanu2019d2}
Mihai Dusmanu, Ignacio Rocco, Tomas Pajdla, Marc Pollefeys, Josef Sivic,
  Akihiko Torii, and Torsten Sattler.
\newblock D2-net: {A} trainable {CNN} for joint description and detection of
  local features.
\newblock In {\em CVPR}, 2019.

\bibitem{fischler1981random}
Martin~A. Fischler and Robert~C. Bolles.
\newblock Random sample consensus: {A} paradigm for model fitting with
  applications to image analysis and automated cartography.
\newblock {\em Commun. ACM}, 1981.

\bibitem{forster2015imu}
Christian Forster, Luca Carlone, Frank Dellaert, and Davide Scaramuzza.
\newblock {IMU} preintegration on manifold for efficient visual-inertial
  maximum-a-posterio estimation.
\newblock In {\em Robotics: Science and Systems}, 2015.

\bibitem{fragoso2020gdls}
Victor Fragoso, Joseph DeGol, and Gang Hua.
\newblock gdls*: Generalized pose-and-scale estimation given scale and gravity
  priors.
\newblock In {\em CVPR}, 2020.

\bibitem{ge2020self}
Yixiao Ge, Haibo Wang, Feng Zhu, Rui Zhao, and Hongsheng Li.
\newblock Self-supervising fine-grained region similarities for large-scale
  image localization.
\newblock In {\em ECCV}, 2020.

\bibitem{germain2020s2dnet}
Hugo Germain, Guillaume Bourmaud, and Vincent Lepetit.
\newblock {S2DNet}: Learning accurate correspondences for sparse-to-dense
  feature matching.
\newblock In {\em ECCV}, 2020.

\bibitem{gordo2017end}
Albert Gordo, Jon Almazan, Jerome Revaud, and Diane Larlus.
\newblock End-to-end learning of deep visual representations for image
  retrieval.
\newblock {\em IJCV}, 2017.

\bibitem{haralick1994review}
Bert~M. Haralick, Chung-Nan Lee, Karsten Ottenberg, and Michael N{\"o}lle.
\newblock Review and analysis of solutions of the three point perspective pose
  estimation problem.
\newblock {\em IJCV}, 1994.

\bibitem{He2016DeepRL}
Kaiming He, Xiangyu Zhang, Shaoqing Ren, and Jian Sun.
\newblock Deep residual learning for image recognition.
\newblock In {\em CVPR}, 2016.

\bibitem{he2022onepose}
Xingyi He, Jiaming Sun, Yuang Wang, Di Huang, Hujun Bao, and Xiaowei Zhou.
\newblock Onepose++: Keypoint-free one-shot object pose estimation without
  {CAD} models.
\newblock In {\em NeurIPS}, 2022.

\bibitem{jeong2019complex}
Jinyong Jeong, Younggun Cho, Young-Sik Shin, Hyunchul Roh, and Ayoung Kim.
\newblock Complex urban dataset with multi-level sensors from highly diverse
  urban environments.
\newblock {\em IJRR}, 2019.

\bibitem{jin2021image}
Yuhe Jin, Dmytro Mishkin, Anastasiia Mishchuk, Jiri Matas, Pascal Fua,
  Kwang~Moo Yi, and Eduard Trulls.
\newblock Image matching across wide baselines: From paper to practice.
\newblock {\em IJCV}, 2021.

\bibitem{Katharopoulos2020TransformersAR}
Angelos Katharopoulos, Apoorv Vyas, Nikolaos Pappas, and Fran{\c{c}}ois
  Fleuret.
\newblock Transformers are rnns: Fast autoregressive transformers with linear
  attention.
\newblock In {\em ICML}, 2020.

\bibitem{kendall2015posenet}
Alex Kendall, Matthew Grimes, and Roberto Cipolla.
\newblock Posenet: {A} convolutional network for real-time 6-dof camera
  relocalization.
\newblock In {\em ICCV}, 2015.

\bibitem{kim2014high}
Hojun Kim, Kyoungah Choi, and Impyeong Lee.
\newblock High accurate affordable car navigation using built-in sensory data
  and images acquired from a front view camera.
\newblock In {\em IV}, 2014.

\bibitem{kneip2011novel}
Laurent Kneip, Davide Scaramuzza, and Roland Siegwart.
\newblock A novel parametrization of the perspective-three-point problem for a
  direct computation of absolute camera position and orientation.
\newblock In {\em CVPR}, 2011.

\bibitem{kukelova2010closed}
Zuzana Kukelova, Martin Bujnak, and Tomas Pajdla.
\newblock Closed-form solutions to minimal absolute pose problems with known
  vertical direction.
\newblock In {\em ACCV}, 2010.

\bibitem{Lepetit2008EPnPAA}
Vincent Lepetit, Francesc Moreno{-}Noguer, and Pascal Fua.
\newblock Ep\emph{n}p: An accurate \emph{O}(\emph{n}) solution to the
  p\emph{n}p problem.
\newblock {\em IJCV}, 2008.

\bibitem{li2018integration}
Yicheng Li, Zhaozheng Hu, Yuezhi Hu, and Duanfeng Chu.
\newblock Integration of vision and topological self-localization for
  intelligent vehicles.
\newblock {\em Mechatronics}, 2018.

\bibitem{li2018megadepth}
Zhengqi Li and Noah Snavely.
\newblock Megadepth: Learning single-view depth prediction from internet
  photos.
\newblock In {\em CVPR}, 2018.

\bibitem{lim2012real}
Hyon Lim, Sudipta~N. Sinha, Michael~F. Cohen, and Matthew Uyttendaele.
\newblock Real-time image-based 6-dof localization in large-scale environments.
\newblock In {\em CVPR}, 2012.

\bibitem{linFocalLossDense2018}
Tsung{-}Yi Lin, Priya Goyal, Ross~B. Girshick, Kaiming He, and Piotr
  Doll{\'{a}}r.
\newblock Focal loss for dense object detection.
\newblock In {\em ICCV}, 2017.

\bibitem{liu2020lsfb}
Haomin Liu, Mingxuan Jiang, Zhuang Zhang, Xiaopeng Huang, Linsheng Zhao, Meng
  Hang, Youji Feng, Hujun Bao, and Guofeng Zhang.
\newblock {LSFB:} {A} low-cost and scalable framework for building large-scale
  localization benchmark.
\newblock In {\em ISMAR}, 2020.

\bibitem{lowe2004distinctive}
David~G. Lowe.
\newblock Distinctive image features from scale-invariant keypoints.
\newblock {\em IJCV}, 2004.

\bibitem{lynen2015get}
Simon Lynen, Torsten Sattler, Michael Bosse, Joel~A. Hesch, Marc Pollefeys, and
  Roland Siegwart.
\newblock Get out of my lab: Large-scale, real-time visual-inertial
  localization.
\newblock In {\em Robotics: Science and Systems}, 2015.

\bibitem{lynen2020large}
Simon Lynen, Bernhard Zeisl, Dror Aiger, Michael Bosse, Joel Hesch, Marc
  Pollefeys, Roland Siegwart, and Torsten Sattler.
\newblock Large-scale, real-time visual-inertial localization revisited.
\newblock {\em IJRR}, 2020.

\bibitem{pinkus1999approximation}
Allan Pinkus.
\newblock Approximation theory of the {MLP} model in neural networks.
\newblock {\em Acta numerica}, 1999.

\bibitem{pion2020benchmarking}
No{\'e} Pion, Martin Humenberger, Gabriela Csurka, Yohann Cabon, and Torsten
  Sattler.
\newblock Benchmarking image retrieval for visual localization.
\newblock In {\em 3DV}, 2020.

\bibitem{qin2019general}
Tong Qin, Shaozu Cao, Jie Pan, and Shaojie Shen.
\newblock A general optimization-based framework for global pose estimation
  with multiple sensors.
\newblock {\em arXiv preprint arXiv:1901.03642}, 2019.

\bibitem{revaud2019learning}
Jerome Revaud, Jon Almaz{\'a}n, Rafael~S. Rezende, and Cesar Roberto~de Souza.
\newblock Learning with average precision: Training image retrieval with a
  listwise loss.
\newblock In {\em ICCV}, 2019.

\bibitem{rusinkiewicz2001efficient}
Szymon Rusinkiewicz and Marc Levoy.
\newblock Efficient variants of the {ICP} algorithm.
\newblock In {\em 3DIM}, 2001.

\bibitem{sarlin2022lamar}
Paul{-}Edouard Sarlin, Mihai Dusmanu, Johannes~L. Sch{\"{o}}nberger, Pablo
  Speciale, Lukas Gruber, Viktor Larsson, Ondrej Miksik, and Marc Pollefeys.
\newblock Lamar: Benchmarking localization and mapping for augmented reality.
\newblock In {\em ECCV}, 2022.

\bibitem{sarlin2021back}
Paul{-}Edouard Sarlin, Ajaykumar Unagar, M{\aa}ns Larsson, Hugo Germain, Carl
  Toft, Viktor Larsson, Marc Pollefeys, Vincent Lepetit, Lars Hammarstrand,
  Fredrik Kahl, and Torsten Sattler.
\newblock Back to the feature: Learning robust camera localization from pixels
  to pose.
\newblock In {\em CVPR}, 2021.

\bibitem{sarlin2019coarse}
Paul-Edouard Sarlin, Cesar Cadena, Roland Siegwart, and Marcin Dymczyk.
\newblock From coarse to fine: Robust hierarchical localization at large scale.
\newblock In {\em CVPR}, 2019.

\bibitem{sarlin2020superglue}
Paul-Edouard Sarlin, Daniel DeTone, Tomasz Malisiewicz, and Andrew Rabinovich.
\newblock Superglue: Learning feature matching with graph neural networks.
\newblock In {\em CVPR}, 2020.

\bibitem{sattler2016efficient}
Torsten Sattler, Bastian Leibe, and Leif Kobbelt.
\newblock Efficient \& effective prioritized matching for large-scale
  image-based localization.
\newblock {\em T-PAMI}, 2016.

\bibitem{sattler2018benchmarking}
Torsten Sattler, Will Maddern, Carl Toft, Akihiko Torii, Lars Hammarstrand,
  Erik Stenborg, Daniel Safari, Masatoshi Okutomi, Marc Pollefeys, Josef Sivic,
  Fredrik Kahl, and Tom{\'{a}}s Pajdla.
\newblock Benchmarking 6dof outdoor visual localization in changing conditions.
\newblock In {\em CVPR}, 2018.

\bibitem{schonberger2016structure}
Johannes~L. Sch{\"{o}}nberger and Jan{-}Michael Frahm.
\newblock Structure-from-motion revisited.
\newblock In {\em CVPR}, 2016.

\bibitem{schops2017multi}
Thomas Sch{\"{o}}ps, Johannes~L. Sch{\"{o}}nberger, Silvano Galliani, Torsten
  Sattler, Konrad Schindler, Marc Pollefeys, and Andreas Geiger.
\newblock A multi-view stereo benchmark with high-resolution images and
  multi-camera videos.
\newblock In {\em CVPR}, 2017.

\bibitem{schreiber2016vehicle}
Markus Schreiber, Hendrik K{\"o}nigshof, Andr{\'e}-Marcel Hellmund, and
  Christoph Stiller.
\newblock Vehicle localization with tightly coupled {GNSS} and visual odometry.
\newblock In {\em IV}, 2016.

\bibitem{shi2012gps}
Yun Shi, Shunping Ji, Zhongchao Shi, Yulin Duan, and Ryosuke Shibasaki.
\newblock Gps-supported visual {SLAM} with a rigorous sensor model for a
  panoramic camera in outdoor environments.
\newblock {\em Sensors}, 2012.

\bibitem{shotton2013scene}
Jamie Shotton, Ben Glocker, Christopher Zach, Shahram Izadi, Antonio Criminisi,
  and Andrew~W. Fitzgibbon.
\newblock Scene coordinate regression forests for camera relocalization in
  {RGB-D} images.
\newblock In {\em CVPR}, 2013.

\bibitem{sivic2003video}
Josef Sivic and Andrew Zisserman.
\newblock Video google: {A} text retrieval approach to object matching in
  videos.
\newblock In {\em ICCV}, 2003.

\bibitem{sun2021loftr}
Jiaming Sun, Zehong Shen, Yuang Wang, Hujun Bao, and Xiaowei Zhou.
\newblock Loftr: Detector-free local feature matching with transformers.
\newblock In {\em CVPR}, 2021.

\bibitem{sun2022onepose}
Jiaming Sun, Zihao Wang, Siyu Zhang, Xingyi He, Hongcheng Zhao, Guofeng Zhang,
  and Xiaowei Zhou.
\newblock Onepose: One-shot object pose estimation without {CAD} models.
\newblock In {\em CVPR}, 2022.

\bibitem{sun2017dataset}
Xun Sun, Yuanfan Xie, Pei Luo, and Liang Wang.
\newblock A dataset for benchmarking image-based localization.
\newblock In {\em CVPR}, 2017.

\bibitem{svarm2016city}
Linus Sv{\"a}rm, Olof Enqvist, Fredrik Kahl, and Magnus Oskarsson.
\newblock City-scale localization for cameras with known vertical direction.
\newblock {\em T-PAMI}, 2016.

\bibitem{svarm2014accurate}
Linus Svarm, Olof Enqvist, Magnus Oskarsson, and Fredrik Kahl.
\newblock Accurate localization and pose estimation for large 3d models.
\newblock In {\em CVPR}, 2014.

\bibitem{sweeney2015efficient}
Chris Sweeney, John Flynn, Benjamin Nuernberger, Matthew Turk, and Tobias
  H{\"o}llerer.
\newblock Efficient computation of absolute pose for gravity-aware augmented
  reality.
\newblock In {\em ISMAR}, 2015.

\bibitem{sweeney2014gdls}
Chris Sweeney, Victor Fragoso, Tobias H{\"o}llerer, and Matthew Turk.
\newblock gdls: A scalable solution to the generalized pose and scale problem.
\newblock In {\em ECCV}, 2014.

\bibitem{taira2018inloc}
Hajime Taira, Masatoshi Okutomi, Torsten Sattler, Mircea Cimpoi, Marc
  Pollefeys, Josef Sivic, Tomas Pajdla, and Akihiko Torii.
\newblock Inloc: Indoor visual localization with dense matching and view
  synthesis.
\newblock In {\em CVPR}, 2018.

\bibitem{tian2019sosnet}
Yurun Tian, Xin Yu, Bin Fan, Fuchao Wu, Huub Heijnen, and Vassileios Balntas.
\newblock Sosnet: Second order similarity regularization for local descriptor
  learning.
\newblock In {\em CVPR}, 2019.

\bibitem{toft2020long}
Carl Toft, Will Maddern, Akihiko Torii, Lars Hammarstrand, Erik Stenborg,
  Daniel Safari, Masatoshi Okutomi, Marc Pollefeys, Josef Sivic, Tomas Pajdla,
  et~al.
\newblock Long-term visual localization revisited.
\newblock {\em T-PAMI}, 2020.

\bibitem{tradacete2018positioning}
Miguel Tradacete, {\'A}lvaro S{\'a}ez, Juan~Felipe Arango, Carlos
  G{\'o}mez~Hu{\'e}lamo, Pedro Revenga, Rafael Barea, Elena
  L{\'o}pez-Guill{\'e}n, and Luis~Miguel Bergasa.
\newblock Positioning system for an electric autonomous vehicle based on the
  fusion of multi-gnss {RTK} and odometry by using an extented kalman filter.
\newblock In {\em WAF}, 2018.

\bibitem{triggs1999bundle}
Bill Triggs, Philip~F. McLauchlan, Richard~I. Hartley, and Andrew~W.
  Fitzgibbon.
\newblock Bundle adjustment - {A} modern synthesis.
\newblock In {\em ICCV}, 1999.

\bibitem{vishal2015accurate}
Kumar Vishal, C.~V. Jawahar, and Visesh Chari.
\newblock Accurate localization by fusing images and gps signals.
\newblock In {\em CVPRW}, 2015.

\bibitem{vysotska2015efficient}
Olga Vysotska, Tayyab Naseer, Luciano Spinello, Wolfram Burgard, and Cyrill
  Stachniss.
\newblock Efficient and effective matching of image sequences under substantial
  appearance changes exploiting {GPS} priors.
\newblock In {\em ICRA}, 2015.

\bibitem{yan2021image}
Shen Yan, Maojun Zhang, Shiming Lai, Yu Liu, and Yang Peng.
\newblock Image retrieval for structure-from-motion via graph convolutional
  network.
\newblock {\em Inf. Sci.}, 2021.

\bibitem{yu2019gps}
Yang Yu, Wenliang Gao, Chengju Liu, Shaojie Shen, and Ming Liu.
\newblock A gps-aided omnidirectional visual-inertial state estimator in
  ubiquitous environments.
\newblock In {\em IROS}, 2019.

\bibitem{zhang2021reference}
Zichao Zhang, Torsten Sattler, and Davide Scaramuzza.
\newblock Reference pose generation for long-term visual localization via
  learned features and view synthesis.
\newblock {\em IJCV}, 2021.

\end{thebibliography}
    }
    
    \end{document}